\documentclass[10pt,twocolumn,letterpaper]{article}

\usepackage{iccv}
\usepackage{times}
\usepackage{epsfig}
\usepackage{graphicx}
\usepackage{amsmath}
\usepackage{amssymb}

\usepackage[pagebackref=true,breaklinks=true,letterpaper=true,colorlinks,bookmarks=false]{hyperref}

\iccvfinalcopy 

\ificcvfinal\fi

\begin{document}

\title{A Video Is Worth Three Views:
Trigeminal Transformers for Video-based Person Re-identification}

\author{
Xuehu Liu$^{\dagger\S}$\quad
Pingping Zhang$^{\dagger\ddagger\S}$\quad
Chenyang Yu$^{\dagger\S}$\quad
Huchuan Lu$^{\dagger\S}$\quad
Xuesheng Qian$^{\pounds}$\quad
Xiaoyun Yang$^{\star}$\\
$^{\dagger}$School of Information and Communication Engineering, Dalian University of Technology,\\
$^{\S}$Ningbo Institute, Dalian University of Technology,\\
$^{\ddagger}$School of Artificial Intelligence, Dalian University of Technology,\\
$^{\pounds}$China Science IntelliCloud Technology Co.Ltd,\\
$^{\star}$Remark Holdings\\
}
\maketitle
\ificcvfinal\thispagestyle{empty}\fi

\begin{abstract}
   Video-based person re-identification (Re-ID) aims to retrieve video sequences of the same person under non-overlapping cameras.
   Previous methods usually focus on limited views, such as spatial, temporal or spatial-temporal view, which lack of the observations in different feature domains.
   To capture richer perceptions and extract more comprehensive video representations, in this paper we propose a novel framework named \textbf{Trigeminal Transformers (TMT)} for video-based person Re-ID.
   More specifically, we design a \textbf{trigeminal feature extractor} to jointly transform raw video data into spatial, temporal and spatial-temporal domain.
   Besides, inspired by the great success of vision transformer, we introduce the transformer structure for video-based person Re-ID.
   In our work, three \textbf{self-view transformers} are proposed to exploit the relationships between local features for information enhancement in spatial, temporal and spatial-temporal domains.
   Moreover, a \textbf{cross-view transformer} is proposed to aggregate the multi-view features for comprehensive video representations.
   The experimental results indicate that our approach can achieve better performance than other state-of-the-art approaches on public Re-ID benchmarks.
   We will release the code for model reproduction.

\end{abstract}

\section{Introduction}
Person re-identification (Re-ID) aims to retrieve given pedestrians across different times and places.
Recently, due to the needs of safe communities, intelligent surveillance and criminal investigations, this task has become a hot research topic.
Meanwhile, since the widespread deployment of video surveillance in cities and the massive data of pedestrians in videos, researchers are paying more and more attention to video-based person Re-ID.
Similar to image-based person Re-ID, video-based person Re-ID also faces many challenges, such as illumination changes, view point difference, complicated backgrounds and person occlusions.
%
%
Different from image-based person Re-ID, video-based person Re-ID utilizes multiple image frames as inputs rather than single image, which will contain additional motion cues, pose variations and multi-view observations.
Although those information is conducive to pedestrian recognition, it also brings more noises and misalignments.
Thus, how to fully utilize the abundant information in sequences is worthy of research in video-based person Re-ID.
%
\begin{figure}
\centering
\resizebox{0.5\textwidth}{!}
{
\begin{tabular}{@{}c@{}c@{}}
\includegraphics[width=0.6\linewidth,height=0.52\linewidth]{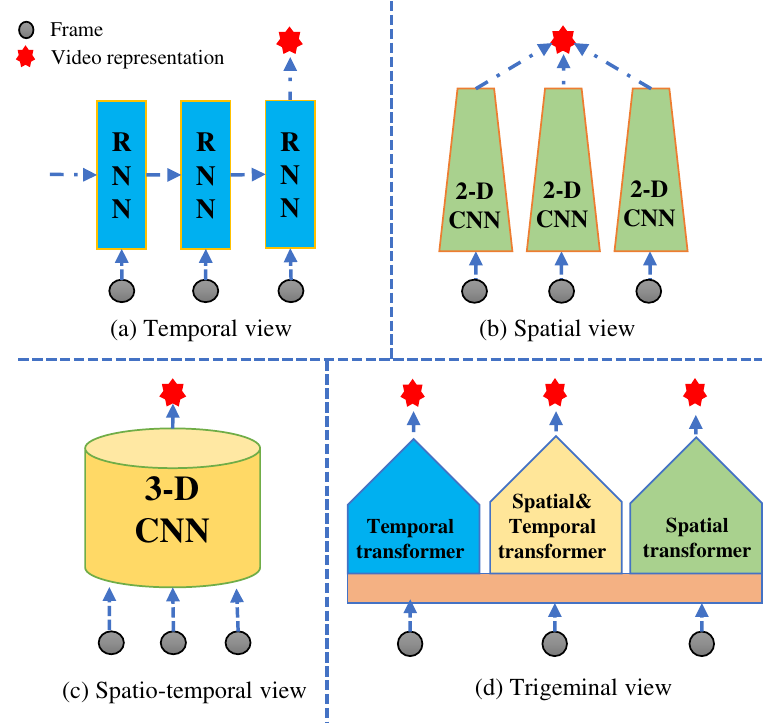} \\
\end{tabular}
}
\vspace{-2mm}
\caption{The (a) (b) (c) shows the previous works with single-view. The (d) shows the trigeminal views in our work.
}
\label{fig:introduction}
\vspace{-6mm}
\end{figure}

In video-based person Re-ID, researchers have explored various methods to process video data.
Some of previous methods utilize spatial feature extractors to obtain attentive feature maps from spatial view.
Li \emph{et al.}~\cite{li2018diversity} design a set of diverse spatial attention modules to extract aligned local features across multiple images.
Meanwhile, temporal learning networks are usually designed for discovering the temporal relationships in temporal domain, such as Recurrent Neural Network (RNN), Long Short Term Memory (LSTM).
For example, Mclaughlin~\emph{et al.}~\cite{mclaughlin2016recurrent} introduce a recurrent architecture to model temporal cues cross frames.
Liu \emph{et al.}~\cite{liu2019spatial} propose a refining recurrent unit to integrate video representation.
Besides, for spatial-temporal observation,
Li~\emph{et al.}~\cite{li2019multi} construct a multi-scale 3D network to learn the multi-scale spatial-temporal cues in video sequences.
However, previous methods often focus on single views, which lacks of the multi-view observations from different view domains.
There are a thousand Hamlets in a thousand people's eyes.
After observing from different perspectives to video data, the model could get more comprehensive and robust video representation.
In our work, we attempt to capture three different observations from spatial, temporal and spatial-temporal domains simultaneously.
The intuitive comparisons are shown in Fig.~\ref{fig:introduction}.
%
%
%
%
%
%
%
%
%

Recently, transformers~\cite{vaswani2017attention,devlin2018bert,brown2020language} have shown the strong representation ability and gained a great success in Natural Language Processing (NLP).
Nowadays, researchers extend transformers for numerous computer-vision applications.
The transformers in vision are mainly developing in two directions: pure-transformer methods~\cite{dosovitskiy2020image,he2021transreid} and ``CNN + transformer" methods~\cite{sun2020transtrack,carion2020end}.
Although pure-transformer methods show great potentiality and are seen as an alternative of CNNs, it is limited by the need of amounts of data for pre-training.
Thus, it is difficult to deploy quickly on a variety of visual tasks.
The ``CNN + transformer" methods retain the powerful spatial feature extraction capability of CNNs, at the same time, introduce the transformers to model the relationships of local features in high-dimensional space.
It becomes a popular paradigm for computer vision.
Inspired by this,
in this work, we combine CNNs and transformers for video-based person Re-ID.
Actually, the interactions in transformers will help to assign different attention weights to local features in spatial-temporal for better video representations.

In this paper, we proposed a novel Trigeminal Transformer (TMT) for video-based person Re-ID.
Our method attempts to capture three observations from spatial, temporal and spatial-temporal domains, simultaneously.
Then, in each view domain, we explore the relationships among local features.
Meanwhile, in order to aggregate multi-view cues and enhance the final representation, the cross-view interactions are taken into consideration in our work.
Specifically, our framework mainly consists of three key modules.
Firstly, we design a trigeminal feature extractor to obtain three different features under spatial, temporal and spatial-temporal views, simultaneously.
Secondly, we introduce three independent transformers as self-view transformers for different views.
Each transformer takes the coarse features as inputs and exploits the relationships between local features for information enhancement.
Finally, we propose a novel cross-view transformer, which models the interactions among features of different views and aggregates them for the final video representation.
Based on above modules, our TMT can not only capture different perceptions in spatial, temporal and spatial-temporal view domains, but also fully aggregate multi-view information to generate more comprehensive video representations.
Extensive experiments on public benchmarks demonstrate that our approach outperforms several state-of-the-art methods.

In summary, our contributions are four folds:
\begin{itemize}
\item
We propose a novel Trigeminal Transformer (TMT) framework for video-based person Re-ID.
\item
We design a trigeminal feature extractor to transform raw video data into spatial, temporal and spatial-temporal view domains for different observations.
\item
We introduce transformer structures for video-based person Re-ID.
Specifically, we propose a self-view transformer to refine single-view features and a cross-view transformer to aggregate multi-view features.
\item
Extensive experiments on public benchmarks demonstrate that our framework synthetically attains a better performance than several state-of-the-art methods.
\end{itemize}

\begin{figure*}
\centering
\resizebox{1.0\textwidth}{!}
{
\begin{tabular}{@{}c@{}c@{}}
\includegraphics[width=1.0\linewidth,height=0.42\linewidth]{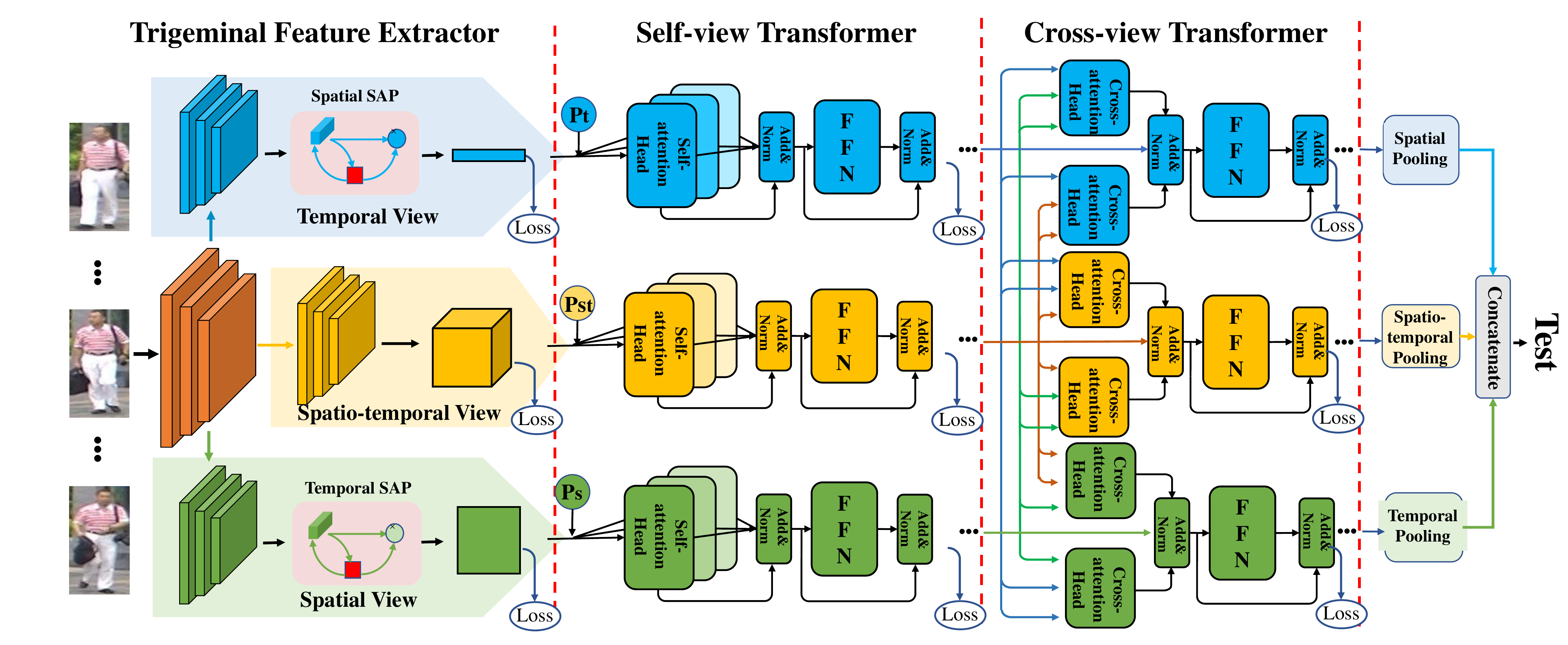} \\
\end{tabular}
}
\caption{The overall structure of our proposed method.
Firstly, the trigeminal feature extractor is utilized to transform raw video data into different view domains individually.
%
Then, the self-view transformers are introduced for feature enhancement.
After that, a cross-view transformer is designed to model the interactions between multiple views for comprehensive observations.
%
}
\label{fig:Framework}
\vspace{-4mm}
\end{figure*}
\section{Related works}
\subsection{Video-based person re-identification}
In recent years, with the rise of deep learning~\cite{deng2009imagenet,ioffe2015batch}, person Re-ID has got a great success and the recognition accuracy has been improved significantly.
Recently, due to the urgent need for video matching in the intelligent community system, video-based person Re-ID has drawn more and more researchers' interest.
Compared with static images, videos contain more views which are worth of observing, such as spatial view, temporal view and spatial-temporal view.
Thus, in video-based person Re-ID, some existing works~\cite{li2018diversity,zhang2020multi, fu2019sta,zhao2019attribute} concentrate on extracting attentive spatial features in the spatial view.
Meanwhile, some methods~\cite{mclaughlin2016recurrent, dai2019video,liu2019spatial, hou2020temporal} attempt to obtain temporal observations by temporal learning mechanisms.
Besides, some approaches~\cite{liu2019dense, li2019multi, wu20193, gu2020appearance} utilize 3D-CNN to jointly explore spatial-temporal cues.
%
%
%
For example, for spatial information,
Li \emph{et al.}~\cite{li2018diversity} attempt to extract extract aligned spatial features across multiple images by diverse spatial attention modules.
Zhao \emph{et al.}~\cite{zhao2019attribute} disentangle frame-wise features for various attribute-aware representations in spatial.
%
%
%
%
%
For temporal information,
Mclaughlin \emph{et al.}~\cite{mclaughlin2016recurrent} utilize recurrent neural networks cross frames for temporal learning.
Liu \emph{et al.}~\cite{liu2019spatial} propose a refining recurrent unit to integrate temporal cues frame by frame.
For spatial-temporal information,
Gu~\emph{et al.}~\cite{gu2020appearance} propose an appearance preserving 3D convolutional network to address appearance destruction and model temporal information.
Meanwhile, beyond single view,
%
Li~\emph{et al.}~\cite{li2019multi} design a two-stream convolutional network to explicitly leverage spatial and temporal cues.
Different from previous methods, in this paper, we design a trigeminal network to transform raw video data into spatial, temporal and spatial-temporal feature space in three different views.
Besides, the self-view transformer and the cross-view transformer are proposed to fully explore discriminative cues in self-view domain and aggregate diverse observations cross multi-view domains for final comprehensive and robust representations.

\subsection{Transformer in vision}
Transformer~\cite{vaswani2017attention} is initially proposed for NLP tasks and brings significant improvement in many tasks~\cite{devlin2018bert,brown2020language}.
Inspired by the powerful ability of transformer for handling sequential data, the transformer-based vision models are springing up like mushrooms, showing the great potential in vision fields.
Researchers have extended transformer for computer vision tasks, such as image and video classification~\cite{girdhar2019video,dosovitskiy2020image,chen2020generative}, object detection~\cite{carion2020end}, semantic segmentation~\cite{wang2020end}, video inpainting~\cite{zeng2020learning} and so on.
For example, Girdhar~\emph{et al.}~\cite{girdhar2019video} introduce an action transformer model to recognize and localize human behaviors in videos.
Carion~\emph{et al.}~\cite{carion2020end} utilize transformers to redesign an end-to-end object detector.
Dosovitskiy~\emph{et al.}~\cite{dosovitskiy2020image} propose the Vision Transformer (ViT) , which applies transformer directly to sequences of image patches and achieves promising results.
Recently, He~\emph{et al.}~\cite{he2021transreid} utilize a pure transformer with a jigsaw patch module for image-based Re-ID.
%
Compared with existing transformers in vision, our transformers are constructed in spatial, temporal and spatial-temporal domains for video-based Re-ID.
%
%
Besides, we propose a novel cross-view transformer to aggregate multi-view cues for comprehensive video representations.
\section{Proposed method}
In this section, we introduce the proposed Trigeminal Transformers (TMT).
%
We first give an overview of the proposed TMT.
Then, we elaborate the key modules in the following subsections.
\subsection{Overview}
The TMT is shown in Fig.~\ref{fig:Framework}.
The overall framework mainly consists of three key modules: Trigeminal Feature Extractor, Self-view Transformer and Cross-view Transformer.
To begin with, we adopt the Restricted Random Sampling (RRS)~\cite{li2018diversity} to generate sequential frames as inputs.
Then, we use the designed trigeminal feature extractor to transform the raw video data into different high-dimensional spaces, which consists of three non-shared embedding branches for multi-view observations.
For extracted initial features in each single-view domain, we utilize the self-view transformer to explore the relationships between local features for information enhancement.
Afterwards, the cross-view transformer is proposed to capture the interaction among multi-view features and adaptively aggregate them for comprehensive representations.
Finally, to train our model, we introduce the Online Instance Matching (OIM)~\cite{xiao2017joint} loss and verification loss.
In the test stage, the feature vectors of different views are concatenated for the retrieval list.

\subsection{Trigeminal feature extractor}
Multi-stream networks are usually used to handle sequential data in action recognition and video classification~\cite{li2019multi,yang2020spatial,simonyan2014two}.
%
%
%
Actually, the success of multi-stream networks can be attributed to the assemble multiple observations from different views, which help to extract more comprehensive representations.
Inspired by this, in our work, we design a trigeminal feature extractor for spatial, temporal and spatial-temporal views.

The structure of our proposed trigeminal feature extractor is shown in the left of Fig.~\ref{fig:Framework}.
Formally, given a long sequence, we firstly sample $T$ frames $\{\textbf{I}^1, \textbf{I}^2, ..., \textbf{I}^T \}$ as the inputs of our network. $T$ is the length of the sequence.
In our work, the ResNet-50~\cite{he2016deep} is used as the basic backbone for frame-wise feature maps.
Different from the previous multi-steam works, we deploy the parameter-shared shallow residual blocks from conv1 to conv3$\_$x in ResNet-50 for reducing network parameters.
Three non-shared conv4$\_$x are deployed after basic CNN as the embedding network and output three video feature cubes, $\textbf{X}^s, \textbf{X}^{t}, \textbf{X}^{st}$, which have same sizes of $T\times H\times W\times C$. $H, W, C$ represents the height, the width and the number of channels, respectively.
%
%
To further separate the feature cubes in high-dimensional space,
we propose a self-attention pooling to project the features into spatial and temporal domains for different views.
%
\begin{figure}
\centering
\resizebox{0.5\textwidth}{!}
{
\begin{tabular}{@{}c@{}c@{}}
\includegraphics[width=1.0\linewidth,height=0.8\linewidth]{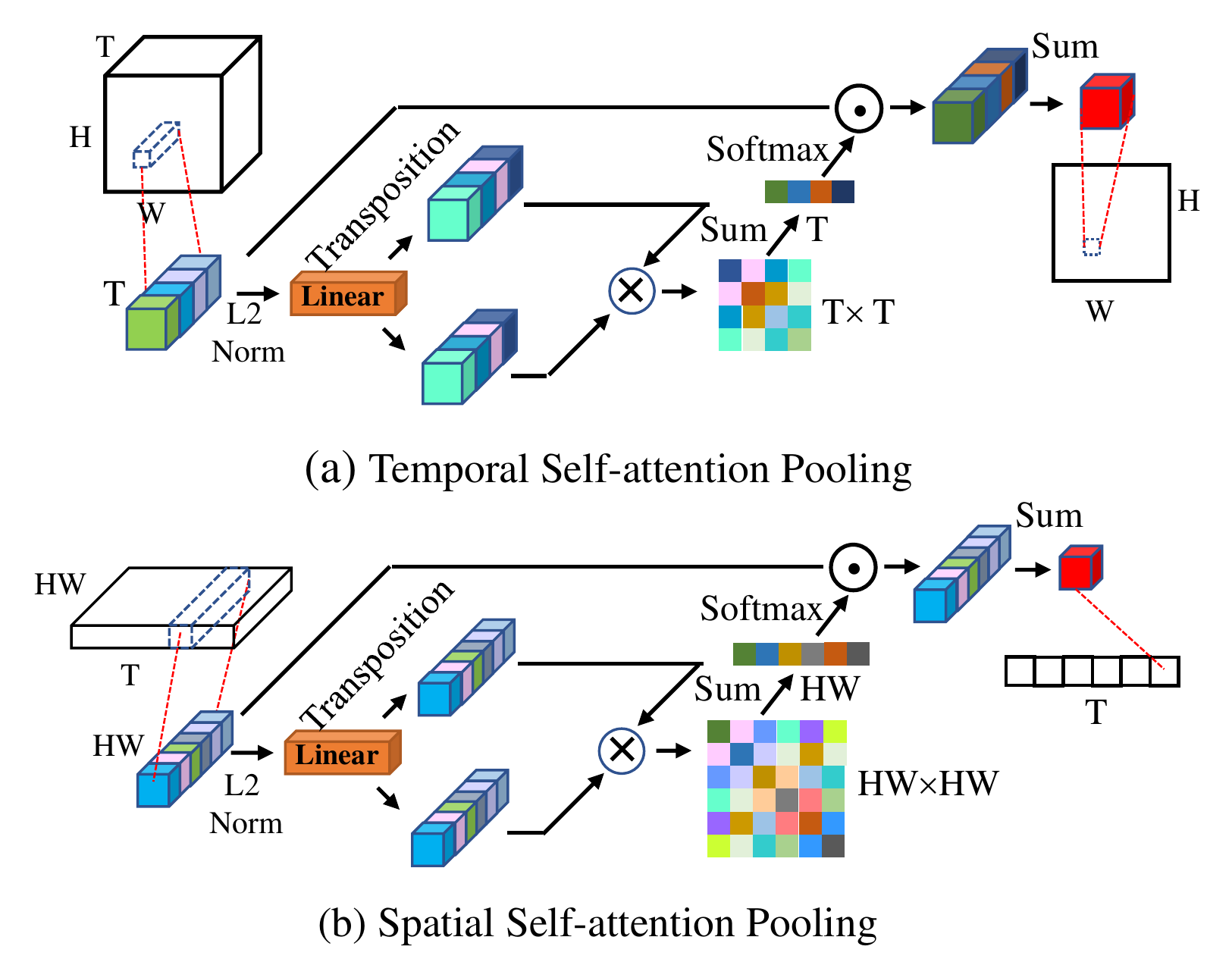} \\
\end{tabular}
}
\vspace{2mm}
\caption{The temporal self-attention pooling (a) and the spatial self-attention pooling (b).
}
\label{fig:SelfAttentionPooling}
\vspace{-2mm}
\end{figure}
\begin{figure*}
\centering
\resizebox{1.0\textwidth}{!}
{
\begin{tabular}{@{}c@{}c@{}}
\includegraphics[width=1.0\linewidth,height=0.5\linewidth]{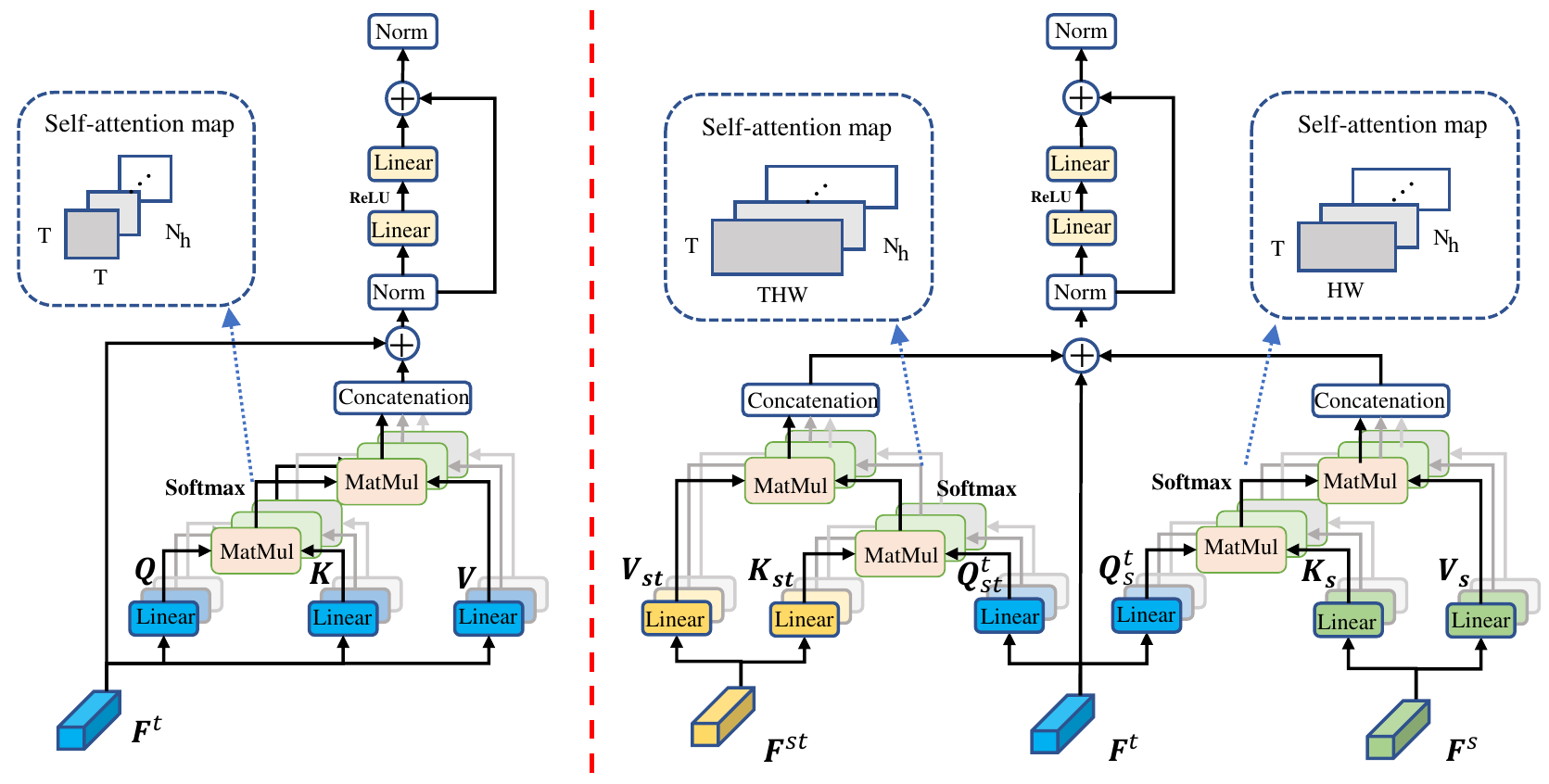} \\
\end{tabular}
}
\vspace{-2mm}
\caption{The self-view transformer (Left) and the cross-view transformer (Right) in temporal domain.
}
\label{fig:transformer}
\vspace{-2mm}
\end{figure*}

\textbf{Self-attention pooling.}
The spatial self-attention pooling and temporal self-attention pooling have the similar structures and are shown in Fig.~\ref{fig:SelfAttentionPooling}.
Given a feature cube $\textbf{X}^s\in {\Re}^{T\times HW\times C} $, we utilize the temporal self-attention pooling to project $\textbf{X}^s$ into spatial domain.
%
To begin with, a linear projection is applied to each spatial local feature $\textbf{X}^{s}_{i}$ $\in {\Re}^{T\times C}$, $i \in [1,H\times W]$ and generates $F_i$ $\in {\Re}^{T\times C}$ by
\begin{equation}\label{attention map}
F_i= \textbf{W}\textbf{X}^s_i
\end{equation}
where $\textbf{W}$ is the parameter of the linear projection.
Then, we apply a matrix multiplication between $F_i$ and its transposition to generate the self-attention matrix $\textbf{M}_i$ $\in {\Re}^{T\times T}$.
\begin{equation}\label{linear projection}
{\textbf{M}}_{i} = F_i {F_i}^T,
\end{equation}
where $(\cdot)^T$ indicates the transpose operation.
After that, we sum the self-attention matrix along one dimension and then perform a softmax operation along other dimension to infer the temporal attention vector $a_i \in {\Re}^T$.
\begin{equation}\label{attention map}
\vspace{-2mm}
a_i= \text{Softmax} (\sum^T_{j=1}\textbf{M}^{j}_i)
\end{equation}
where $j \in [1, T]$.
Then, a dot production is applied between each local spatial feature $\textbf{X}^{s}_{i}$ and its temporal attention vector $a_i$.
By this way, we obtain the attentive spatial feature $\hat{\textbf{X}}^s_i \in {\Re}^{T\times C}$
\begin{equation}\label{multiapplication}
\hat{\textbf{X}}^s_i = \textbf{X}^s_{i} \odot a_{i}
\end{equation}
\vspace{-2mm}
\vspace{-2mm}
\begin{equation}\label{multiapplication}
g_i = \sum^T_t \textbf{X}^s_{i,t}
\end{equation}
where $\odot$ represent the matrix dot multiplication and $g_i \in {\Re}^{C}$.
Finally, the local spatial feature sets $\textbf{F}^s = \{g_1,..., g_i,..., g_{H\times W}\}$ ($i\in [1, H\times W]$) as the output of temporal self-attention pooling .

%
The resulting spatial features can be regard as the weighted feature in temporal, which utilizes the relationships cross frames.
By this way, we can pool the feature cube $\textbf{X}^s\in {\Re}^{T\times HW\times C}$ to $\textbf{F}^s\in {\Re}^{HW\times C} $.
Noted that, the spatial self-attention pooling has the similar process as the temporal self-attention pooling.
The difference is the size of the self attention matrices.
In the spatial self-attention pooling,
the size of the spatial attention matrix is $HW\times HW$, and the output can be represented by $\textbf{F}^t\in {\Re}^{T\times C}$.
%
%
Thus, our trigeminal feature extractor utilizes a partially shared network to project raw video data into independent spatial, temporal and spatial-temporal domains for multi-view observations.
After that, self-view transformers and a cross-view transformer are deployed for further global feature enhancement.

\subsection{Self-view transformer}
Based on the above trigeminal feature extractor, we can easily obtain particular features in different views.
However, these features lack global observations in each self-view domain.
Recently, transformer models have been proposed to explore the interactions among contextual information and deliver a great success.
Actually, the interactions among global features are beneficial to mine more discriminative cues.
Inspired by the strong capacity of transformer and its great potentiality in vision fields, in our work, we introduce the transformer structure into different view domains for feature enhancement.
Self-view transformers are multi-layer architectures formed by stacking blocks on top of one another.
Each block of transformer is composed of a multi-head self-attention layer, a position-wise feed-forward network, layer normalization modules and residual connections.
For the temporal domain, the obtained feature $\textbf{F}^t \in {\Re}^{T\times C}$ from the trigeminal feature extractor is passed into the self-view transformer (The left part of Fig.~\ref{fig:transformer}), which can be formulated as follows.
First, a positional encoding is added to the original feature for the positional information on temporal, which can be trainable like~\cite{dosovitskiy2020image}.
Then, the feature with position is passed through the multi-head self-attention layer.
In each head, the feature is firstly feed into three linear transformations to generate feature $Q$, $K$ and $V$, where $Q,K,V\in {\Re}^{T\times d}$. $d= \frac{C}{N_h}$ and $N_h$ is the number of heads.
The self attention operation is defined as:
\vspace{-1mm}
\begin{equation}\label{selfattention}
A_h = \text{Softmax}(\frac{\textbf{Q}\textbf{K}^T}{\sqrt{d}})\textbf{V},
\end{equation}
\vspace{-1mm}
The outputs of multiple heads, $A_1,\cdots, A_{N_h}$, are concatenated
together as the output of the multi-headed self-attention layer.
The input and output of the multi-head self-attention layer are connected by residual connections and a normalization layer,
\vspace{-1mm}
\begin{equation}\label{add&norm}
\textbf{F}^t = \text{LayerNorm}(\textbf{F}^t + \text{MultiHead}(\textbf{F}^t)).
\end{equation}
\vspace{-1mm}
The position-wise feed-forward network (FFN) is applied after the multi-head self-attention layer.
Specifically, the FFN consists of two linear transformation layers and a ReLU activation function within them, which can be denoted as the following function:
\begin{equation}\label{FFN}
\textbf{F}^t = \textbf{W}_2 \sigma (\textbf{W}_1 \textbf{F}^t),
\end{equation}
where $\textbf{W}_1$ and $\textbf{W}_2$ are the parameters of two linear transformation layers and $\sigma$ is the ReLU activation function.

In the spatial and spatial-temporal views, the $\textbf{F}^s$ and $\textbf{F}^{st}$ obtained from the trigeminal feature extractor are passed through the spatial transformer and spatial-temporal transformer individually, which have the same structures as the temporal transformer.
In our work, we deploy the transformers in each self-view domain, called self-view transformers, to model the relationships among local features for further feature enhancement.
\subsection{Cross-view transformer}
In each domain, self-view transformers help to refine the initial feature by exploiting the attention relationships in current view.
Actually, the attention relationships cross multiple views are also instructive to the feature refinement.
Considering this, in this paper, a cross-view transformer is constructed cross different view domains, which has spatial, temporal and spatial-temporal views simultaneously.
The right part of Fig.~\ref{fig:transformer} shows the detailed structure of the proposed cross-view transformer, which consists of multi-head cross-attention layer, domain-wise feed-forward networks, layer normalization modules and residual connections.
%
%
Formally, in the temporal-based cross-attention head, six linear projections are applied to generate six features $\textbf{Q}^t_s,\textbf{Q}^t_{st},\textbf{K}_s, \textbf{K}_{st}, \textbf{V}_s, \textbf{V}_{st}$,
where $\textbf{Q}^t_s, \textbf{Q}^t_{st}$ $\in {\Re}^{T\times d}$ are from $\textbf{F}^t$, $\textbf{K}_s, \textbf{V}_s$ $\in {\Re}^{HW\times d}$ are from $\textbf{F}^s$ and $\textbf{K}_{st},\textbf{V}_{st}$ $\in {\Re}^{THW\times d}$ are from $\textbf{F}^{st}$.
$d= \frac{C}{N_h}$ and $N_h$ is the number of heads.
In one head, the cross-attention is formulated as:
\begin{equation}\label{selfattention}
\hat{\textbf{F}}^t_s = \text{Softmax}(\frac{\textbf{Q}^t_s \textbf{K}_s^T}{\sqrt{d}}) \textbf{V}_s,
\end{equation}
\begin{equation}\label{selfattention}
\hat{\textbf{F}}^t_{st} = \text{Softmax}(\frac{\textbf{Q}^t_{st} \textbf{K}^T_{st}}{\sqrt{d}}) \textbf{V}_{st},
\end{equation}
where $(\cdot)^T$ is the transposition.
The concatenation of multi-head outputs can be represented by $\textbf{F}^t_{s}$ and $\textbf{F}^t_{st}$, which  have the same size to $\textbf{F}^t$.
Then, the output of multi-head cross-attention head can be expressed as:
\begin{equation}\label{add&norm}
\textbf{F}^t = \text{LayerNorm}(\textbf{F}^t + \textbf{F}^t_s +  \textbf{F}^t_{st}).
\end{equation}
After that, similar to self-view transformer, a domain-wise feed-forward network is applied.
\begin{equation}\label{FFN}
\textbf{F}^t = \textbf{W}_4 \sigma (\textbf{W}_3 \textbf{F}^t),
\end{equation}
where $\textbf{W}_3$ and $\textbf{W}_4$ are parameters of the two linear transformation layers and $\sigma$ represents the ReLU activation function.
By this way, in the temporal domain, its feature can aggregate the related information from other views.
For the spatial and spatial-temporal views, the interactions are explored in similar cross-view transformers.
Finally, the features from different views are adaptively  integrated by proposed cross-view transformer and generate more comprehensive representations.
\begin{table*}
\caption{Ablation results of key components on MARS, iLIDS-VID and PRID2011.}
\label{table: ablation of key components}
\begin{center}
\doublerulesep=0.1pt
\resizebox{1.0\textwidth}{!}
{
\begin{tabular}{|c|c|c|c|c|c|c|c|c|c|c|c|c|c|c|c|c|c|c|c|c|c|c|c|c|c|c|c|c|c|c|c|c|c|c|c|c|c|c|c|c|c|c|c|c|c|c|c|c|c|c|c|c|c|c|c|c|c|c|c|c|c|c|c|c|c|c|c|c|c|c|c|c|c|c|c|c|c|c|c|c|c|c|c|c|c|c|c|c|c|c|c|c|c|c|c|c|c|c|c|c|c|c|c|c|c|c|c|c|c|c|c|c|c|}
\hline
\multicolumn{20}{l}{}
&\multicolumn{16}{|c|}{MARS}
&\multicolumn{12}{|c}{iLIDS-VID}
&\multicolumn{12}{|c}{PRID2011}
\\
\hline
\hline
\multicolumn{20}{l|}{Method}
&\multicolumn{4}{c}{mAP}&\multicolumn{4}{c}{Rank-1}
&\multicolumn{4}{c}{Rank-5}&\multicolumn{4}{c}{Rank-20}
&\multicolumn{4}{|c}{Rank-1}
&\multicolumn{4}{c}{Rank-5}&\multicolumn{4}{c}{Rank-20}
&\multicolumn{4}{|c}{Rank-1}
&\multicolumn{4}{c}{Rank-5}&\multicolumn{4}{c}{Rank-20}
\\
\hline
\hline
\multicolumn{20}{l|}{Baseline}
&\multicolumn{4}{c}{81.2}&\multicolumn{4}{c}{88.5}
&\multicolumn{4}{c}{95.5}&\multicolumn{4}{c}{97.9}
&\multicolumn{4}{|c}{88.2}
&\multicolumn{4}{c}{97.6}&\multicolumn{4}{c}{99.7}
&\multicolumn{4}{|c}{93.5}
&\multicolumn{4}{c}{98.7}&\multicolumn{4}{c}{99.7}
\\
\hline
\hline
\multicolumn{20}{l|}{+ Three branches}
&\multicolumn{4}{c}{82.0}&\multicolumn{4}{c}{88.9}
&\multicolumn{4}{c}{95.6}&\multicolumn{4}{c}{98.0}
&\multicolumn{4}{|c}{89.3}
&\multicolumn{4}{c}{97.8}&\multicolumn{4}{c}{99.6}
&\multicolumn{4}{|c}{94.4}
&\multicolumn{4}{c}{99.1}&\multicolumn{4}{c}{99.8}
\\
\hline
\hline
\multicolumn{20}{l|}{+ Trigeminal feature extractor}
&\multicolumn{4}{c}{83.9}&\multicolumn{4}{c}{89.7}
&\multicolumn{4}{c}{96.1}&\multicolumn{4}{c}{98.2}
&\multicolumn{4}{|c}{89.7}
&\multicolumn{4}{c}{98.4}&\multicolumn{4}{c}{99.7}
&\multicolumn{4}{|c}{94.8}
&\multicolumn{4}{c}{99.1}&\multicolumn{4}{c}{100}
\\
\multicolumn{4}{l}{}&\multicolumn{16}{l|}{Spatial view}
&\multicolumn{4}{c}{80.2}&\multicolumn{4}{c}{86.9}
&\multicolumn{4}{c}{95.2}&\multicolumn{4}{c}{97.5}
&\multicolumn{4}{|c}{86.7}
&\multicolumn{4}{c}{97.7}&\multicolumn{4}{c}{99.5}
&\multicolumn{4}{|c}{91.3}
&\multicolumn{4}{c}{98.2}&\multicolumn{4}{c}{99.8}
\\
\multicolumn{4}{l}{}&\multicolumn{16}{l|}{Temporal view}
&\multicolumn{4}{c}{82.8}&\multicolumn{4}{c}{89.3}
&\multicolumn{4}{c}{95.8}&\multicolumn{4}{c}{98.1}
&\multicolumn{4}{|c}{86.3}
&\multicolumn{4}{c}{98.0}&\multicolumn{4}{c}{99.8}
&\multicolumn{4}{|c}{93.2}
&\multicolumn{4}{c}{98.7}&\multicolumn{4}{c}{100}
\\
\multicolumn{4}{l}{}&\multicolumn{16}{l|}{Spatial-temporal view}
&\multicolumn{4}{c}{83.4}&\multicolumn{4}{c}{89.5}
&\multicolumn{4}{c}{95.7}&\multicolumn{4}{c}{98.1}
&\multicolumn{4}{|c}{87.6}
&\multicolumn{4}{c}{97.9}&\multicolumn{4}{c}{99.6}
&\multicolumn{4}{|c}{92.7}
&\multicolumn{4}{c}{98.6}&\multicolumn{4}{c}{99.8}
\\
\hline
\hline
\multicolumn{20}{l|}{+ Self-view transformer}
&\multicolumn{4}{c}{85.6}&\multicolumn{4}{c}{90.8}
&\multicolumn{4}{c}{96.7}&\multicolumn{4}{c}{98.4}
&\multicolumn{4}{|c}{90.6}
&\multicolumn{4}{c}{98.2}&\multicolumn{4}{c}{100}
&\multicolumn{4}{|c}{95.7}
&\multicolumn{4}{c}{99.0}&\multicolumn{4}{c}{99.8}
\\
\multicolumn{4}{l}{}&\multicolumn{16}{l|}{Spatial view}
&\multicolumn{4}{c}{83.4}&\multicolumn{4}{c}{90.0}
&\multicolumn{4}{c}{96.1}&\multicolumn{4}{c}{98.2}
&\multicolumn{4}{|c}{88.9}
&\multicolumn{4}{c}{97.8}&\multicolumn{4}{c}{99.7}
&\multicolumn{4}{|c}{93.3}
&\multicolumn{4}{c}{98.5}&\multicolumn{4}{c}{99.8}
\\
\multicolumn{4}{l}{}&\multicolumn{16}{l|}{Temporal view}
&\multicolumn{4}{c}{83.5}&\multicolumn{4}{c}{89.6}
&\multicolumn{4}{c}{95.7}&\multicolumn{4}{c}{98.0}
&\multicolumn{4}{|c}{88.4}
&\multicolumn{4}{c}{97.5}&\multicolumn{4}{c}{99.8}
&\multicolumn{4}{|c}{94.7}
&\multicolumn{4}{c}{98.8}&\multicolumn{4}{c}{100}
\\
\multicolumn{4}{l}{}&\multicolumn{16}{l|}{Spatial-temporal view}
&\multicolumn{4}{c}{84.6}&\multicolumn{4}{c}{90.2}
&\multicolumn{4}{c}{96.6}&\multicolumn{4}{c}{98.2}
&\multicolumn{4}{|c}{90.2}
&\multicolumn{4}{c}{98.2}&\multicolumn{4}{c}{99.9}
&\multicolumn{4}{|c}{93.7}
&\multicolumn{4}{c}{98.7}&\multicolumn{4}{c}{99.6}
\\
\hline
\hline
\multicolumn{20}{l|}{+ Cross-view transformer}
&\multicolumn{4}{c}{85.8}&\multicolumn{4}{c}{91.2}
&\multicolumn{4}{c}{97.3}&\multicolumn{4}{c}{98.8}
&\multicolumn{4}{|c}{91.3}
&\multicolumn{4}{c}{98.6}&\multicolumn{4}{c}{100}
&\multicolumn{4}{|c}{96.4}
&\multicolumn{4}{c}{99.3}&\multicolumn{4}{c}{100}
\\
\hline
\hline
\end{tabular}
}
\vspace{-8mm}
\end{center}
\end{table*}
\begin{figure*}
\centering
\resizebox{1.0\textwidth}{!}
{
\begin{tabular}{@{}c@{}c@{}}
\includegraphics[width=1.0\linewidth,height=0.25\linewidth]{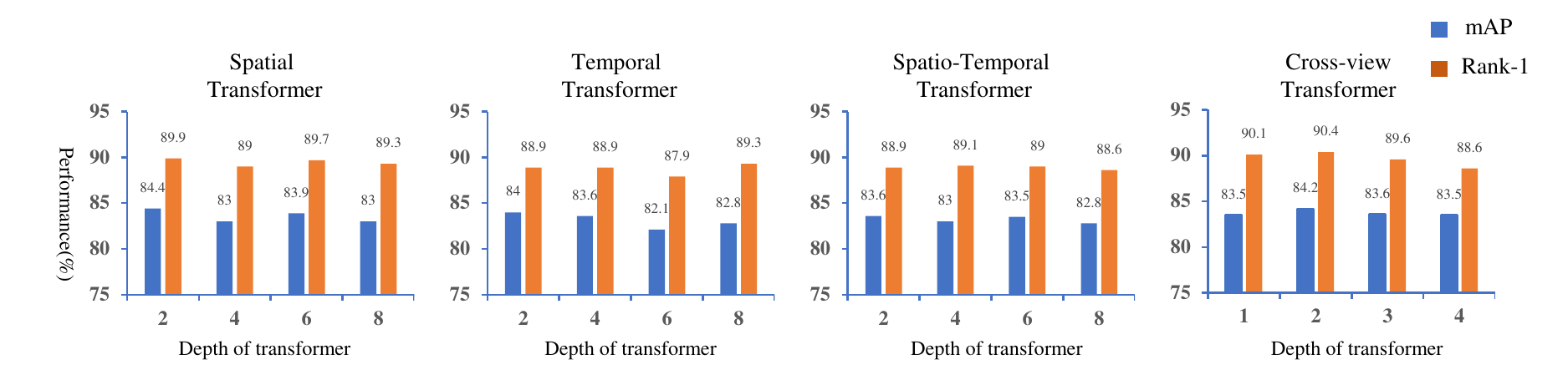} \\
\end{tabular}
}
\vspace{-2mm}
\caption{Ablation results on the depth of self-view transformers and the cross-view transformer in on MARS.
}
\label{fig:ablation on the depth of transformer}
\vspace{-2mm}
\end{figure*}
\begin{table}
\caption{Ablation results on the length of sequences on MARS.}
\vspace{-4mm}
\label{table: ablation the length of sequence}
\begin{center}
\doublerulesep=0.1pt
\resizebox{0.5\textwidth}{!}
{
\begin{tabular}{|c|c|c|c|c|c|c|c|c|c|c|c|c|c|c|c|c|c|c|c|c|c|c|c|c|c|c|c|c|c|c|c|c|c|c|c|c|c|c|c|}
\hline
\multicolumn{8}{l|}{Methods}
&\multicolumn{12}{c}{Temporal}&\multicolumn{12}{|c}{Spatial-temporal}
\\
\multicolumn{8}{l|}{}
&\multicolumn{12}{c}{Transformer}&\multicolumn{12}{|c}{Transformer}
\\
\hline
\hline
\multicolumn{8}{l|}{Length (T)}
&\multicolumn{4}{c}{mAP}&\multicolumn{4}{c}{Rank-1}
&\multicolumn{4}{c}{Rank-20}
&\multicolumn{4}{|c}{mAP}&\multicolumn{4}{c}{Rank-1}
&\multicolumn{4}{c}{Rank-20}
\\
\hline
\hline
\multicolumn{8}{l|}{6}
&\multicolumn{4}{c}{83.4}&\multicolumn{4}{c}{88.6}&\multicolumn{4}{c}{98.0}
&\multicolumn{4}{|c}{82.6}&\multicolumn{4}{c}{89.5}&\multicolumn{4}{c}{97.8}
\\
\multicolumn{8}{l|}{8}
&\multicolumn{4}{c}{84.0}&\multicolumn{4}{c}{88.7}&\multicolumn{4}{c}{98.1}
&\multicolumn{4}{|c}{84.0}&\multicolumn{4}{c}{89.8}&\multicolumn{4}{c}{98.5}
\\
\multicolumn{8}{l|}{10}
&\multicolumn{4}{c}{83.2}&\multicolumn{4}{c}{89.1}&\multicolumn{4}{c}{97.8}
&\multicolumn{4}{|c}{83.8}&\multicolumn{4}{c}{89.2}&\multicolumn{4}{c}{97.9}
\\
\multicolumn{8}{l|}{12}
&\multicolumn{4}{c}{83.0}&\multicolumn{4}{c}{88.6}&\multicolumn{4}{c}{97.9}
&\multicolumn{4}{|c}{83.7}&\multicolumn{4}{c}{88.7}&\multicolumn{4}{c}{98.0}
\\
\hline
\hline
\end{tabular}
}
\vspace{-4mm}
\end{center}
\end{table}
\begin{table}
\caption{Ablation results on the size of spatial feature maps on MARS.}
\vspace{-6mm}
\label{table: ablation the size of spatial feature map}
\begin{center}
\doublerulesep=0.1pt
\resizebox{0.5\textwidth}{!}
{
\begin{tabular}{|c|c|c|c|c|c|c|c|c|c|c|c|c|c|c|c|c|c|c|c|c|c|c|c|c|c|c|c|c|c|c|c|c|c|c|c|c|c|c|c|}
\hline
\multicolumn{8}{l|}{Methods}
&\multicolumn{12}{c}{Spatial}&\multicolumn{12}{|c}{Spatial-temporal}
\\
\multicolumn{8}{l|}{}
&\multicolumn{12}{c}{Transformer}&\multicolumn{12}{|c}{Transformer}
\\
\hline
\hline
\multicolumn{8}{l|}{Size (H$\times$W)}
&\multicolumn{4}{c}{mAP}&\multicolumn{4}{c}{Rank-1}&\multicolumn{4}{c}{Rank-20}
&\multicolumn{4}{|c}{mAP}&\multicolumn{4}{c}{Rank-1}&\multicolumn{4}{c}{Rank-20}
\\
\hline
\hline
\multicolumn{8}{l|}{8$\times$4}
&\multicolumn{4}{c}{82.4}&\multicolumn{4}{c}{89.2}&\multicolumn{4}{c}{98.0}
&\multicolumn{4}{|c}{83.2}&\multicolumn{4}{c}{90.0}&\multicolumn{4}{c}{97.8}
\\
\multicolumn{8}{l|}{16$\times$8}
&\multicolumn{4}{c}{84.4}&\multicolumn{4}{c}{89.9}&\multicolumn{4}{c}{98.4}
&\multicolumn{4}{|c}{84.0}&\multicolumn{4}{c}{89.8}&\multicolumn{4}{c}{98.5}
\\
\hline
\hline
\end{tabular}
}
\vspace{-6mm}
\end{center}
\end{table}

\section{Experiments}
\subsection{Datasets and evaluation protocols}
In this paper,  we adopt three widely-used benchmarks to evaluate our proposed method, \emph{i.e.}, iLIDS-VID~\cite{wang2014person}, PRID-2011~\cite{hirzer2011person} and MARS~\cite{zheng2016mars}.
iLIDS-VID~\cite{wang2014person} and PRID-2011~\cite{hirzer2011person} are two small datasets.
%
Both of them collected images by two cameras.
iLIDS-VID~\cite{wang2014person} has 600 video sequences of 300 different identities.
PRID-2011~\cite{hirzer2011person} consists of 400 image sequences for 200 identities from two non-overlapping cameras.
%
%
%
MARS~\cite{zheng2016mars} is one of large-scale datasets, and consists of 1,261 identities around 18,000 video sequences.
All the video sequences are captured by at least 2 cameras.
Noted that, there are round 3,200 distractors sequences in the dataset to  simulate actual detect conditions.
For evaluation, the Cumulative Matching Characteristic (CMC) table and mean Average Precision (mAP) are adopted following previous works for MARS dataset. For more details, we refer readers to the original paper~\cite{zheng2016person}.
In terms of iLIDS-VID and PRID2011,
there is one single correct match in the gallery set.
Thus, only the cumulative re-identification accuracy is reported.
\subsection{Implementation details}
We implement our framework based on the Pytorch~\footnote{https://pytorch.org/} toolbox.
The experimental devices include an Intel i4790 CPU and two NVIDIA 3090 (24G memory).
%
%
Experimentally, we set the $batchsize$ = 16.
In our work, if not specified, we set the length of sequence $T$ = 8 and the depth of transformer to 2.
Each image in a sequence is resized to 256$\times$128 and augmented by random cropping, horizontal flipping and random erasing.
The ResNet-50~\cite{he2016deep} pre-trained on the ImageNet dataset~\cite{deng2009imagenet} is used as our backbone network.
Following previous works~\cite{sun2018beyond}, we remove the last spatial down-sampling operation to increase the feature resolution.
%
%
%
%
In a mini-batch, we selected a number of positive and negative sequence pairs for verification loss.
Besides, we utilize frame-wise and video-wise OIM losses to supervise the whole network following~\cite{chen2018video}.
In experiments, the outputs of backbone in three branches are also supervised to strength the learning of the underlying network.
During training, we train our network for 50 epochs and the learning rate is decayed by 10 at every 15 epochs.
The whole network is updated by stochastic gradient descent~\cite{bottou2010large} algorithm with an initial learning rate of $10^{-3}$, weight decay of $5\times10^{-4}$ and nesterov momentum of 0.9.
We will release the source code for model reproduction.
%
\begin{table*}
\begin{center}
\doublerulesep=0.5pt
\caption{Comparison with state-of-the-art video-based person re-identification methods on MARS, iLIDS-VID and PRID2011.}
\label{table:soa method}
\vspace{2mm}
\resizebox{0.9\textwidth}{!}
{
\begin{tabular}{|c|c|c|c|c|c|c|c|c|c|c|c|c|c|c|c|c|c|c|c|c|c|c|c|c|c|c|c|c|c|c|c|c|c|c|c|c|c|c|c|c|c|c|c|c|c|c|c|c|c|c|c|c|c|c|c|c|c|c|c|c|c|c|c|c|c|c|c|c|c|c|c|c|c|c|c|c|c|c|c|c|c|c|c}
\hline
\multicolumn{8}{c|}{}
&\multicolumn{8}{c|}{}
&\multicolumn{16}{c|}{MARS}
&\multicolumn{12}{c|}{iLIDS-VID}
&\multicolumn{12}{c}{PRID2011}
\\
\multicolumn{8}{c|}{Methods}
&\multicolumn{8}{c}{Source}
&\multicolumn{4}{|c}{mAP}
&\multicolumn{4}{c}{Rank-1}
&\multicolumn{4}{c}{Rank-5}
&\multicolumn{4}{c|}{Rank-20}
&\multicolumn{4}{c}{Rank-1}
&\multicolumn{4}{c}{Rank-5}
&\multicolumn{4}{c|}{Rank-20}
&\multicolumn{4}{c}{Rank-1}
&\multicolumn{4}{c}{Rank-5}
&\multicolumn{4}{c}{Rank-20}
\\
\hline
\multicolumn{8}{l|}{SeeForest~\cite{zhou2017see}}
&\multicolumn{8}{c|}{CVPR17}
&\multicolumn{4}{c}{50.7}
&\multicolumn{4}{c}{70.6}
&\multicolumn{4}{c}{90.0}
&\multicolumn{4}{c}{97.6}
&\multicolumn{4}{|c}{55.2}
&\multicolumn{4}{c}{86.5}
&\multicolumn{4}{c|}{97.0}
&\multicolumn{4}{c}{79.4}
&\multicolumn{4}{c}{94.4}
&\multicolumn{4}{c}{99.3}
\\
\multicolumn{8}{l|}{ASTPN~\cite{xu2017jointly}}
&\multicolumn{8}{c|}{ICCV17}
&\multicolumn{4}{c}{-}
&\multicolumn{4}{c}{44}
&\multicolumn{4}{c}{70}
&\multicolumn{4}{c}{81}
&\multicolumn{4}{|c}{62}
&\multicolumn{4}{c}{86}
&\multicolumn{4}{c|}{98}
&\multicolumn{4}{c}{77}
&\multicolumn{4}{c}{95}
&\multicolumn{4}{c}{99}
\\
\multicolumn{8}{l|}{Snippet~\cite{chen2018video}}
&\multicolumn{8}{c|}{CVPR18}
&\multicolumn{4}{c}{76.1}
&\multicolumn{4}{c}{86.3}
&\multicolumn{4}{c}{94.7}
&\multicolumn{4}{c}{98.2}
&\multicolumn{4}{|c}{85.4}
&\multicolumn{4}{c}{96.7}
&\multicolumn{4}{c|}{99.5}
&\multicolumn{4}{c}{93.0}
&\multicolumn{4}{c}{99.3}
&\multicolumn{4}{c}{100}
\\
\multicolumn{8}{l|}{STAN~\cite{li2018diversity}}
&\multicolumn{8}{c|}{CVPR18}
&\multicolumn{4}{c}{65.8}
&\multicolumn{4}{c}{82.3}
&\multicolumn{4}{c}{-}
&\multicolumn{4}{c}{-}
&\multicolumn{4}{|c}{80.2}
&\multicolumn{4}{c}{-}
&\multicolumn{4}{c|}{-}
&\multicolumn{4}{c}{93.2}
&\multicolumn{4}{c}{-}
&\multicolumn{4}{c}{-}
\\
\multicolumn{8}{l|}{STMP~\cite{liu2019spatial}}
&\multicolumn{8}{c|}{AAAI19}
&\multicolumn{4}{c}{72.7}
&\multicolumn{4}{c}{84.4}
&\multicolumn{4}{c}{93.2}
&\multicolumn{4}{c}{96.3}
&\multicolumn{4}{|c}{84.3}
&\multicolumn{4}{c}{96.8}
&\multicolumn{4}{c|}{99.5}
&\multicolumn{4}{c}{92.7}
&\multicolumn{4}{c}{98.8}
&\multicolumn{4}{c}{99.8}
\\
\multicolumn{8}{l|}{M3D~\cite{li2019multi}}
&\multicolumn{8}{c|}{AAAI19}
&\multicolumn{4}{c}{74.0}
&\multicolumn{4}{c}{84.3}
&\multicolumn{4}{c}{93.8}
&\multicolumn{4}{c}{97.7}
&\multicolumn{4}{|c}{74.0}
&\multicolumn{4}{c}{94.3}
&\multicolumn{4}{c|}{-}
&\multicolumn{4}{c}{94.4}
&\multicolumn{4}{c}{100}
&\multicolumn{4}{c}{-}
\\
\multicolumn{8}{l|}{Attribute~\cite{zhao2019attribute}}
&\multicolumn{8}{c|}{CVPR19}
&\multicolumn{4}{c}{78.2}
&\multicolumn{4}{c}{87.0}
&\multicolumn{4}{c}{95.4}
&\multicolumn{4}{c}{\underline{98.7}}
&\multicolumn{4}{|c}{86.3}
&\multicolumn{4}{c}{87.4}
&\multicolumn{4}{c|}{99.7}
&\multicolumn{4}{c}{93.9}
&\multicolumn{4}{c}{99.5}
&\multicolumn{4}{c}{100}
\\
\multicolumn{8}{l|}{VRSTC~\cite{hou2019vrstc}}
&\multicolumn{8}{c|}{CVPR19}
&\multicolumn{4}{c}{82.3}
&\multicolumn{4}{c}{88.5}
&\multicolumn{4}{c}{96.5}
&\multicolumn{4}{c}{97.4}
&\multicolumn{4}{|c}{83.4}
&\multicolumn{4}{c}{95.5}
&\multicolumn{4}{c|}{99.5}
&\multicolumn{4}{c}{-}
&\multicolumn{4}{c}{-}
&\multicolumn{4}{c}{-}
\\
\multicolumn{8}{l|}{GLTR~\cite{li2019global}}
&\multicolumn{8}{c|}{ICCV19}
&\multicolumn{4}{c}{78.5}
&\multicolumn{4}{c}{87.0}
&\multicolumn{4}{c}{95.8}
&\multicolumn{4}{c}{98.2}
&\multicolumn{4}{|c}{86.0}
&\multicolumn{4}{c}{98.0}
&\multicolumn{4}{c|}{-}
&\multicolumn{4}{c}{95.5}
&\multicolumn{4}{c}{\textbf{100}}
&\multicolumn{4}{c}{-}
\\
\multicolumn{8}{l|}{COSAM~\cite{subramaniam2019co}}
&\multicolumn{8}{c|}{ICCV19}
&\multicolumn{4}{c}{79.9}
&\multicolumn{4}{c}{84.9}
&\multicolumn{4}{c}{95.5}
&\multicolumn{4}{c}{97.9}
&\multicolumn{4}{|c}{79.6}
&\multicolumn{4}{c}{95.3}
&\multicolumn{4}{c|}{-}
&\multicolumn{4}{c}{-}
&\multicolumn{4}{c}{-}
&\multicolumn{4}{c}{-}
\\
\multicolumn{8}{l|}{MGRA~\cite{zhang2020multi}}
&\multicolumn{8}{c|}{CVPR20}
&\multicolumn{4}{c}{\textbf{85.9}}
&\multicolumn{4}{c}{88.8}
&\multicolumn{4}{c}{\underline{97.0}}
&\multicolumn{4}{c}{98.5}
&\multicolumn{4}{|c}{88.6}
&\multicolumn{4}{c}{98.0}
&\multicolumn{4}{c|}{99.7}
&\multicolumn{4}{c}{95.9}
&\multicolumn{4}{c}{99.7}
&\multicolumn{4}{c}{100}
\\
\multicolumn{8}{l|}{STGCN~\cite{yang2020spatial}}
&\multicolumn{8}{c|}{CVPR20}
&\multicolumn{4}{c}{83.7}
&\multicolumn{4}{c}{89.9}
&\multicolumn{4}{c}{-}
&\multicolumn{4}{c}{-}
&\multicolumn{4}{|c}{-}
&\multicolumn{4}{c}{-}
&\multicolumn{4}{c|}{-}
&\multicolumn{4}{c}{-}
&\multicolumn{4}{c}{-}
&\multicolumn{4}{c}{-}
\\
\multicolumn{8}{l|}{AFA~\cite{chen2020temporal}}
&\multicolumn{8}{c|}{ECCV20}
&\multicolumn{4}{c}{82.9}
&\multicolumn{4}{c}{90.2}
&\multicolumn{4}{c}{96.6}
&\multicolumn{4}{c}{-}
&\multicolumn{4}{|c}{88.5}
&\multicolumn{4}{c}{96.8}
&\multicolumn{4}{c|}{99.7}
&\multicolumn{4}{c}{-}
&\multicolumn{4}{c}{-}
&\multicolumn{4}{c}{-}
\\
\multicolumn{8}{l|}{TCLNet~\cite{hou2020temporal}}
&\multicolumn{8}{c|}{ECCV20}
&\multicolumn{4}{c}{85.1}
&\multicolumn{4}{c}{89.8}
&\multicolumn{4}{c}{-}
&\multicolumn{4}{c}{-}
&\multicolumn{4}{|c}{86.6}
&\multicolumn{4}{c}{-}
&\multicolumn{4}{c|}{-}
&\multicolumn{4}{c}{-}
&\multicolumn{4}{c}{-}
&\multicolumn{4}{c}{-}
\\
\multicolumn{8}{l|}{GRL~\cite{liu2021watching}}
&\multicolumn{8}{c|}{CVPR21}
&\multicolumn{4}{c}{84.8}
&\multicolumn{4}{c}{\underline{91.0}}
&\multicolumn{4}{c}{96.7}
&\multicolumn{4}{c}{98.4}
&\multicolumn{4}{|c}{\underline{90.4}}
&\multicolumn{4}{c}{\underline{98.3}}
&\multicolumn{4}{c|}{\underline{99.8}}
&\multicolumn{4}{c}{\underline{96.2}}
&\multicolumn{4}{c}{\underline{99.7}}
&\multicolumn{4}{c}{\underline{100}}
\\
\hline
\multicolumn{8}{l|}{TMT(Ours)}
&\multicolumn{8}{c|}{-}
&\multicolumn{4}{c}{\underline{85.8}}
&\multicolumn{4}{c}{\textbf{91.2}}
&\multicolumn{4}{c}{\textbf{97.3}}
&\multicolumn{4}{c}{\textbf{98.8}}
&\multicolumn{4}{|c}{\textbf{91.3}}
&\multicolumn{4}{c}{\textbf{98.6}}
&\multicolumn{4}{c|}{\textbf{100}}
&\multicolumn{4}{c}{\textbf{96.4}}
&\multicolumn{4}{c}{99.3}
&\multicolumn{4}{c}{\textbf{100}}
\\
\hline
\end{tabular}
}
\vspace{-6mm}
\end{center}
\end{table*}
\subsection{Ablation study}
To investigate the effectiveness of our TMT, we conduct a series of experiments on three public benchmarks: MARS, iLIDS-VID and PRID2011.

\textbf{Effectiveness of key components}.
We gradually add our modules to backbone for ablation analysis.
The results are shown in Tab.~\ref{table: ablation of key components}.
In this table, ``Baseline'' represents the ResNet-50 as the feature extractor following directly spatial and temporal average pooling for the final video representation.
``+ Three branches'' means that three non-shared Res4$\_$x blocks in ResNet-50 are deployed as same as the trigeminal feature extractor, which brings slight improvements.
Compared with ``+ Three branches'', ``+ Trigeminal feature extractor'' represents the proposed temporal and spatial self-attention pooling replacing the direct average pooling in spatial and temporal branches respectively.
We can see that our proposed self-attention pooling is beneficial to transform same features into different domains and improves $1.9\%$ in term of mAP on MARS.
Moreover, we add three self-view transformers in spatial, temporal and spatial-temporal domains for `` + Self-view transformer''.
The individual view-wise feature attains a significant improvement than ``+ Trigeminal feature extractor'' on three benchmarks.
Compared with ``Baseline'', our proposed three self-view transformers can improve the mAP by $4.4\%$ and the Rank-1 accuracy by $2.3\%$ on MARS.
The improvements indicate that self-view transformers could help to model the relationships among global features for better performances.
Noted that, the combination of three views boosts higher results than single view, which clarifies our statement: a video is worth three views.
Last but not least,  we apply the proposed cross-view transformer in `` + Cross-view transformer''.
The performance has been further improved on three benchmarks.
It gains $0.4\%$, $0.7\%$ and $0.7\%$ in term of the Rank-1 accuracy on MARS, iLIDS-VID and PRID2011 respectively.
It indicates that the cross-view transformer is beneficial to aggregate multi-view observations for comprehensive video representations.

\textbf{Effect of the length of sequences}.
In Tab.~\ref{table: ablation the length of sequence}, the ablation results show the influence of different sequence length on MARS.
The temporal or spatial-temporal transformer is applied to the baseline for single-view observation.
We can see that, the temporal and spatial-temporal transformer are sensitive to the length of the sequences.
When varying the length of the sequence to 8, both of temporal and spatial-temporal transformers get best performance.

\textbf{Effect of the size of spatial feature map}.
We also perform ablation experiments to investigate the effect of varying the spatial size of feature maps.
The standard ResNet-50 is utilized and generates the frame-wise feature maps, which have the spatial size of $8\times 4$.
When we remove the last spatial down-sampling operation of ResNet-50, the size of feature maps will increase to $16\times 8$.
In this way, the proposed the spatial or spatial-temporal transformer could be applied to different features with different sizes.
The ablation results for spatial and spatial-temporal transformer are reported in Tab.~\ref{table: ablation the size of spatial feature map}.
The results show that the increments of spatial size will gain significant improvement.

\textbf{Effect of the depth of transformer}.
The depth of transformer is an important factor for exploring the interaction of contextual information.
The increments of depth will improve the representation capacity of transformer, but it will also bring training difficulties.
In Fig.~\ref{fig:ablation on the depth of transformer}, performances with different depths are reported on different view-wise transformers.
The experiments are conducted on MARS.
We add the spatial, temporal or spatial-temporal transformer to  `` Baseline''  and add the cross-view transformer to  `` + Three branches''  respectively.
From the results, we can see that, the depths of achieving best performance in one transformers are different to other transformers.
Besides, for the cross-view transformers, a two-layer structure gains better than shallower transformers or deeper transformers.
%
\subsection{Compared with the state of the arts}
In this section, our TMT is compared with state-of-the-art methods on MARS, iLIDS-VID and PRID2011.
The results are reported in Tab.~\ref{table:soa method}.
One can observe that, on MARS dataset, our TMT attains comparable even better performances than other compared methods.
In addition, our method achieves highest Rank-1 accuracy on three benchmarks.
Compared with existing methods, our method integrates multi-view cues to obtain more comprehensive video representations.
MGRA~\cite{zhang2020multi} extracts multi-granularity spatial cues under the guidance of a global view, which gains remarkable $85.9\%$ mAP on MARS dataset.
Attribute~\cite{zhao2019attribute} mines various attribute-aware features in spatial for alignment.
Those methods focus on capturing diverse spatial features and gain remarkable performances.
Even so, comapred with MGRA~\cite{zhang2020multi},  our method improves the performances by $\textbf{2.4}\%$ and $\textbf{2.7}\%$ in terms of mAP and Rank-1 accuracy on MARS and iLIDS.
%
%
GLTR~\cite{li2019global} attempts to model the multi-granular temporal dependencies for short and long-term temporal cues.
GRL~\cite{liu2021watching} utilizes a bi-direction temporal learning to refine and accumulate disentangled spatial features.
Those methods are indeed helpful to capture the discriminative cues in temporal for better recognition accuracy.
Especially, GRL~\cite{liu2021watching} attains an expressive Rank-1 accuracy on iLIDS-VID datasets.
Our method has still gained $\textbf{0.9}\%$ improvements about
Rank-1 accuracy than GRL~\cite{liu2021watching}.
Besides, STA~\cite{fu2019sta} introduces a free-parameter spatial-temporal attention module to weight local features in spatial-temporal domain.
Different from the above single-view methods, our proposed TMT resembles spatial, temporal and spatial-temporal observations and achieves better results on three public datasets.
Meanwhile, it is worth noting that some methods construct two-stream networks for different feature representations.
For example, M3D~\cite{li2019multi} combines 3D-CNN and 2D-CNN to explicitly leverage spatial and temporal cues.
STGCN~\cite{yang2020spatial} constructs two parallel graph convolutional networks to explore the relations in spatial and temporal domains.
Compared with these two-stream approaches, our method introduces a self-view transformer into single-view domain for feature enhancement, and a cross-view transformer to aggregate multiple views for better representations.
In this way, our method surpasses STGCN~\cite{yang2020spatial} by $\textbf{2.1}\%$, $\textbf{1.3}\%$ in terms of mAP and Rank-1 accuracy on MARS.
In summary, our method performs better than most existing state-of-the-arts.
These results validate the superiority of our method.

\section{Conclusion}
In this paper, we propose a novel framework named Trigeminal Transformer for video-based person Re-ID.
A trigeminal feature extractor is designed to capture spatial, temporal and spatial-temporal view domains, in which the proposed temporal self-attention pooling and spatial self-attention pooling are applied to transform features into spatial and temporal domains respectively.
Besides, self-view transformers are introduced to explore the relationships among global features for feature enhancement in self-view domains.
In addition, a cross-view transformer is proposed to model the interactions among multi-view features for more comprehensive representations.
Based on our proposed modules, we model a video from spatial, temporal and spatial-temporal views.
%
%
Extensive experiments on there public benchmarks demonstrate that our approach performs better than some state-of-the-arts.
%

{\small
\bibliographystyle{ieee_fullname}
\bibliography{egbib}
}

\end{document}